\documentclass[letterpaper]{article} 
\usepackage{aaai25}  
\usepackage{times}  
\usepackage{helvet}  
\usepackage{courier}  
\usepackage[hyphens]{url}  
\usepackage{graphicx} 
\usepackage{natbib}  
\usepackage{caption} 
\usepackage{algorithm}
\usepackage{algorithmic}

\usepackage{newfloat}
\usepackage{listings}
\DeclareCaptionStyle{ruled}{labelfont=normalfont,labelsep=colon,strut=off} 
\floatstyle{ruled}
\newfloat{listing}{tb}{lst}{}
\floatname{listing}{Listing}
\usepackage{amssymb}
\usepackage{amsmath}
\usepackage{booktabs}
\usepackage{multirow}
\usepackage{subfig}

\title{Why Does Dropping Edges Usually Outperform Adding Edges in Graph Contrastive Learning?}
\author{
    Yanchen Xu\textsuperscript{\rm 1,2}\footnotemark[2],
    Siqi Huang\textsuperscript{\rm 1,2}\footnotemark[2],
    Hongyuan Zhang\textsuperscript{\rm 2,3}\footnotemark[1],
    Xuelong Li\textsuperscript{\rm 2}\footnote{: Corresponding authors. $^\dag$: Equal Contribution.}
}
\affiliations{
    \textsuperscript{\rm 1} School of Artificial Intelligence, OPtics and ElectroNics (iOPEN), Northwestern Polytechnical University\\
    \textsuperscript{\rm 2}Institute of Artificial Intelligence (TeleAI), China Telecom\\
    \textsuperscript{\rm 3}The University of Hong Kong\\
    \{yanchenxu.tj, 4777huang, hyzhang98\}@gmail.com, xuelong\_li@ieee.org
}

\usepackage{bibentry}

\begin{document}

\maketitle

\begin{abstract}
Graph contrastive learning (GCL) has been widely used as an effective self-supervised learning method for graph representation learning.
However, how to apply adequate and stable graph augmentation to generating proper views for contrastive learning remains an essential problem. Dropping edges is a primary augmentation in GCL while adding edges is not a common method due to its unstable performance.
To our best knowledge, there is no theoretical analysis to study why dropping edges usually outperforms adding edges.
To answer this question, we introduce a new metric, namely Error Passing Rate (EPR), to quantify how a graph fits the network.
Inspired by the theoretical conclusions and the idea of positive-incentive noise, we propose a novel GCL algorithm, Error-PAssing-based Graph Contrastive Learning (EPAGCL), which uses both edge adding and edge dropping as its augmentations.
To be specific, we generate views by adding and dropping edges based on the weights derived from EPR.
Extensive experiments on various real-world datasets are conducted to validate the correctness of our theoretical analysis and the effectiveness of our proposed algorithm.
Our code is available at: https://github.com/hyzhang98/EPAGCL.
\end{abstract}
\section{Introduction}
Graph contrastive learning (GCL) has been a hot research topic in graph representation learning \cite{GCA,BGRL,AnchorGAE,SGNN}.
It originates from contrastive learning \cite{SimCLR,MOCO,SimCSE,CoDA,CLIP}, which generates different augmented views and maximizes the similarity between positive pairs
 \cite{InfoNCE}.

In particular, compared with CL, it has been shown that how to generate proper graph augmentation views is a crucial problem in GCL \cite{Infomin, VINCE}.
Due to the particularity of the structure of graph data, the reliable unsupervised data augmentation schemes are greatly limited.
Inspired by Dropout \cite{Dropout}, many researchers chose to randomly mask the features as the attribute augmentations. 
Moreover, for many GCL methods \cite{GRACE,BGRL,GraphCL}, the most popular scheme is to change the graph topology, most of which randomly drops edges to generate views.
Apart from random edge-dropping \cite{DropEdge}, researchers have tried various methods to generate stable views.
MVGRL \cite{MVGRL} introduced diffusion kernels for augmentation.
GCA \cite{GCA} augmented the graph based on the importance of edges.
And GCS \cite{GCS} adaptively screened the semantic-related substructure in graphs for contrastive learning.
\begin{figure}[t]
    \centering
    \subfloat[Performance on WikiCS]{
        \includegraphics[width=0.45\linewidth]{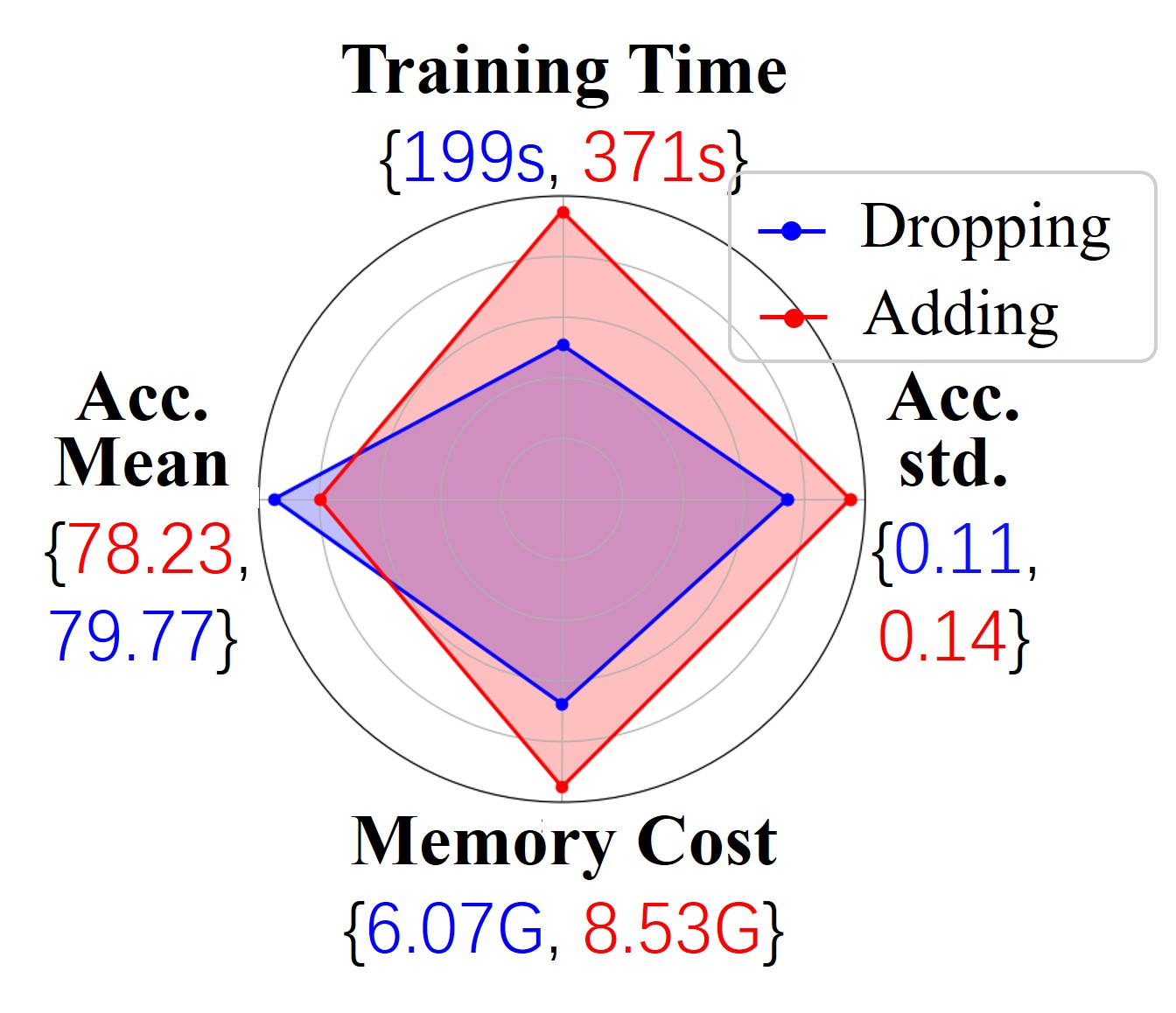}
    }
    \hfill
    \subfloat[Performance on CiteSeer]{
        \includegraphics[width=0.45\linewidth]{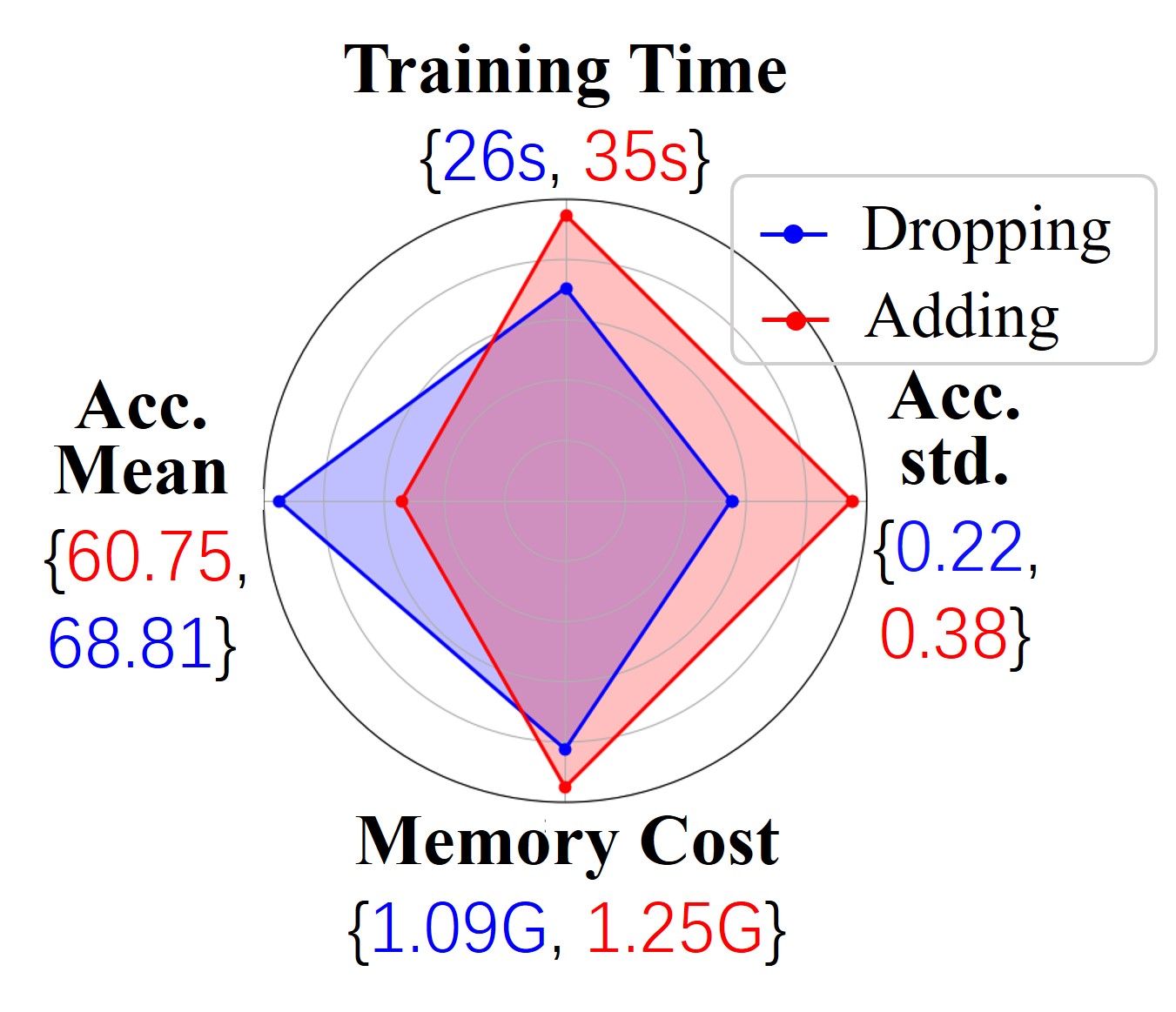}
    }
    \caption{Performance on WikiCS and CiteSeer. All the hyper-parameters are fixed. Edge adding and edge dropping are the only augmentation employed respectively.}
    \label{Radar}
\end{figure}
Although various works have been done to investigate augmentation schemes for GCL, edge adding, as a simple way of graph topology augmentation, is only used in supervised learning \cite{GAUG,AdaEdge}.
\textbf{Why is edge dropping a widely accepted way for GCL other than edge adding?}
Can edge adding be modified to achieve similar effect as edge dropping?
In this paper, we attempt to answer the above questions theoretically.

To begin with, a simple experiment is conducted on WikiCS \cite{WikiCS} and CiteSeer \cite{Planetoid} to compare the two strategies of augmentations.
We use the same hyper-parameters and employ Adding Edges or Dropping Edges as the only augmentation, respectively.
The result is shown in Figure \ref{Radar}.
It is easy to conclude the disadvantage of edge adding in all respects.

Generally speaking, there are two main challenges for edge adding as augmentation:
\begin{itemize}
    \item[$\mathcal{C}1$] For most graphs, the number of edges takes only a small portion of all node pairs, which means that there are much more choices for edge-adding than edge-dropping.
    Hence, the memory and time burden of edge-adding are significant.
    \item[$\mathcal{C}2$] Without labels, we have no idea about the influence of the edge perturbation.
    In that case, a suitable metric is needed to measure it.
\end{itemize}

To deal with the challenges above, we introduce a metric of graph, namely Error Passing Rate (EPR), to quantify the fitness between the graph and Graph Neural Network (GNN).
To be specific, EPR measures \textit{how much message is wrongly passed during the aggregation in GNNs}.
For a certain graph, a lower EPR means that the aggregation is more efficient, which is what we expect.
Consequently, \textit{the aim of augmentation can be converted to generating a view with low EPR}.

In this paper, through mathematical derivation, we prove that \textit{the degree attribute of two nodes can measure the effect on EPR of adding or dropping the edge between them}.
To maintain a relatively low EPR, it is important to ensure that the EPR of the graph will not increase too much even if an edge is wrongly added or dropped.
To this end, for stability, node pairs that correspond to low-level effect will be chosen to add or drop the edges between them, as a solution to $\mathcal{C}2$.
As a result, the nodes can be pre-screened to reduce the memory and time burden, which resolves $\mathcal{C}1$.

Inspired by our theoretical analysis, we propose a novel adaptive augmentation framework for GCL, where a graph is augmented through selective edge perturbation and random feature mask.
For edge perturbation, the possibility that an edge is added to or dropped from the graph is based on the magnitude of its effect on EPR.
Briefly speaking, the possibility is high if the magnitude of effect is relatively low.
This helps to maintain the EPR of augmented views in a low level.
The augmentation is then equipped with an InfoNCE-like objective \cite{InfoNCE,GRACE} for contrastive learning, namely Error-PAssing-rate-based Graph Contrastive Learning (EPAGCL).
\section{Related Works}
\subsection{Graph Contrastive Learning}
Various graph contrastive learning methods have been proposed \cite{TADropEdge,CCA-SSG,GREET} in recent years.
In general, Graph Contrastive Learning (GCL) methods aims to learn representations by contrasting positive and negative samples.

The researchers mainly focus on the contrastive objective.
For the sake of simplicity, random edge dropping and feature masking ars widely used \cite{DropEdge,GRACE,BGRL}.
Furthermore, GraphCL \cite{GraphCL} and InfoGCL \cite{InfoGCL} employ two more augmentations, node dropping and subgraph sampling, which change the node attribute and structure property of the graph at the same time.
There are also some methods using their own augmentation methods.
MVGRL \cite{MVGRL} generates views through graph diffusion.
SimGCL \cite{SimGCL} adds uniform noise to the embedding space to create contrastive views.

\subsection{Adaptive Data Augmentations for Graph}
Some researchers have investigated methods for adaptive data augmentations \cite{pi-noise, PiNDA} for GCL.
JOAO \cite{JOAO} adaptively and dynamically selects data augmentations for specific data.
GCA \cite{GCA} sets the probability of dropping edges and masking features according to their importance.
ADGCL \cite{ADGCL} optimizes adversarial graph augmentation strategies for contrastive learning.
GCS \cite{GCS} screens sub-structure in graphs with the help of a gradient-based graph contrastive saliency.
NCLA \cite{NCLA} learns graph augmentations by multi-head graph attention mechanism.

Our proposed augmentation method is similar to GCA.
However, edges that EPAGCL tends to add or drop are likely to be preserved in GCA.
What's more, our theory is rigorously derived quantitatively, and our augmentation scheme is proved to be more stable.
\subsection{Adding Edges as Augmentation}
Adding edges is also used as augmentation for graph data is some supervised learning methods.
AdaEdge \cite{AdaEdge} and GAUG \cite{GAUG} both add edges based on prediction metrics, which cannot be used in self-supervised learning.

Both methods tend to minimize the negative impact of the augmentation, which is the same as our proposed framework.
To be specific, both of them reduce the probability of error augmentation, while EPAGCL, inspired by pi-noise (a theoretical framework to learn beneficial noise) \cite{pi-noise}, reduces the magnitude of the impact if an error occurs.
\begin{figure*}[t]
\centering
\includegraphics[width=\textwidth]{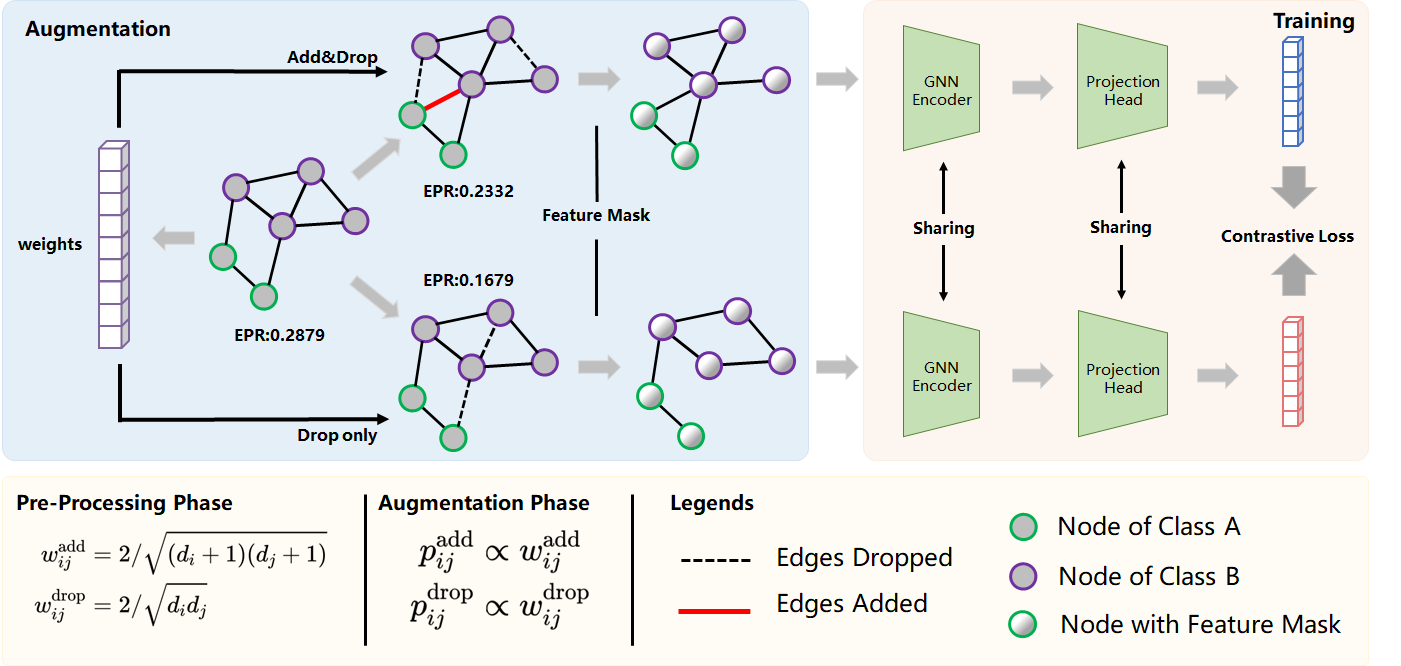}
\caption{Framework of EPAGCL: Before training, the weight of all existing edges and candidate edge for adding is computed according to the graph structure. We then generate two views adaptively based on the weights. Specifically, we add edges to and drop edges from the graph to obtain one view while drop edges only from the graph to obtain another. A random feature mask is then employed. After that, the two views are fed to a shared Graph Neural Network (GNN) with a projection head for representation learning. The model is trained with a contrastive objective.}
\label{Framework}
\end{figure*}
\section{The Proposed Method}
In this section, we first introduce the formal definition of Error Passing Rate (EPR) to quantify how a graph fits the GNN.
Then, the augmentations of graph topology are interpreted based on EPR.
Finally, the Error-PAssing-based Graph Contrastive Learning (EPAGCL) method induced by the theoretical analysis is elaborated.
The Framework is illustrated in Figure \ref{Framework}.
\subsection{Preliminaries}
Let $\mathcal{G} = (\mathcal{V}, \mathcal{E})$ denote an undirected graph without isolated points, whose node set is $\mathcal{V} = \{v_1, v_2, \cdots, v_N\}$ and edge set is $\mathcal{E} \subseteq \mathcal{V} \times \mathcal{V}$.
The adjacency matrix of $\mathcal{G}$ is denoted as $A \in \mathbb{R}^{N\times N}$ with $A_{ij} = 1$ if $(v_i, v_j)\in \mathcal{E}$ and $A_{ij} = 0$ otherwise.
The degree matrix of $\mathcal{G}$, denoted as $D$, is a diagonal matrix with $D_{ii} = \sum_{j}A_{ij}$.
Furthermore, let $\widetilde{D} = D + I$, $\widetilde{A} = A + I$, where $I$ is an $N$-dimensional identity matrix.
$\forall v\in \mathcal{V}$, the set of its neighbours in graph $\mathcal{G}$ is denoted as $N_v$ with $|N_v|=d_v=D_{vv}$.
Let $d_{min}=\min_{v\in V}d_v$, $d_{max}=\max_{v\in V}d_v$.

For $v_i, v_j\in \mathcal{V}\; s.t.\,(v_i, v_j)\notin\mathcal{E}$, suppose that adding edges $(v_i, v_j)$ to graph $\mathcal{G}$ yields graph $\mathcal{G}'=(\mathcal{V}, \mathcal{E}')$, where $\mathcal{E}'=\mathcal{E}\cup \{(v_i, v_j)\}$.
Relatively, $A'$, $D'$, $\widetilde{A}'$, and $\widetilde{D}'$ of graph $\mathcal{G}'$ are defined in the same way as above.

\subsection{Error Passing Rate}
As message passing is a popular paradigm for GNNs, we introduce Error Passing Rate based on message passing mechanisms to measure how the graph fits the network:
\paragraph{Definition 3.1.} For a graph $\mathcal{G}$, the Error Passing Rate (EPR) $r_\mathcal{G}$ denotes the ratio of the amount of message wrongly passed in the network, i.e.,
\begin{equation}
r_\mathcal{G}=M_{wp}/M,\nonumber    
\end{equation}
where $M_{wp}$ is the amount of message wrongly passed, while $M$ is the amount of all the message passed.

Note that $r_\mathcal{G}$ will change with different network structure.
In this paper, Graph Convolutional Networks (GCN) \cite{GCN} are taken as the target.
The feed forward propagation in GCN can be recursivly conducted as
\begin{equation}
    H^{(l+1)}=\sum(\widetilde{D}^{-1/2}\widetilde{A}\widetilde{D}^{-1/2}H^{(l)}W^{(l)}),
    \label{GCN network}
\end{equation}
where $H^{(l+1)} = \{h_1^{(l+1)}, \cdots, h_N^{(l+1)}\}$ is the hidden vector of the $(l+1)$-th layer with $h_i^{(l)}$ as the hidden feature of node $v_i$, and $\sum(\cdot)$ is a nonlinear function. Eq. (\ref{GCN network}) suggests that the topological structure of the graph is represented as $\widetilde{D}^{-1/2}\widetilde{A}\widetilde{D}^{-1/2}\overset{\triangle}{=}\hat{A}$ in the network, corresponding to the message-passing mechanism.
Specifically, the amount of message passed from node $v_i$ to $v_j$ can be measured by $\hat{A}_{ij}$.
Thus, we can obtain the representation of $r_{\mathcal{G}}$.
\paragraph{Proposition 3.2.} Given an undirected graph $\mathcal{G}$, its EPR $r_{\mathcal{G}}$ in GCN can be formulated as
\begin{equation}
    r_{\mathcal{G}}=\sum_{(v_i,v_j)\in E}\hat{A}_{ij}/\sum_{(v_i,v_j)\in \mathcal{E}}\hat{A}_{ij},
    \label{EPR}
\end{equation}
where $E$ denotes the set of edges in $\mathcal{E}$ that links nodes of different, underlying classes which are agnostical during the self-supervised training, namely error edge set.
\subsection{Why is Adding Edges Usually Worse?}
An ordinary idea is to persue a smaller $r_{\mathcal{G}}$.
This naturally leads to a question: \textit{how will $r_{\mathcal{G}}$ change after adding $(v_i, v_j)$ into graph $\mathcal{G}$ as an augmentation?} 

It is complex to calculate the change of $r_\mathcal{G}$ with Eq. (\ref{EPR}).
To simplify the calculation, we group the edges according to their relationship with $v_i$ and $v_j$, between which no edge lies.
To be specific, for graph $\mathcal{G}$, we divide $\mathcal{E}$ into three parts:
\begin{equation}
    \begin{aligned}
        \mathcal{E}_1 &= \{(u,v)|u,v\in V\backslash\{v_i,v_j\}\}\cap\mathcal{E},\\
        \mathcal{E}_2 &= \{(v_i,w)|w\in V\}\cap\mathcal{E},\\
        \mathcal{E}_3 &= \{(v_j,w)|w\in V\}\cap\mathcal{E}.\nonumber
    \end{aligned}    
\end{equation}
Note that $\mathcal{E}_1\cup \mathcal{E}_2\cup \mathcal{E}_3 = \mathcal{E}$.
Similarly, the error edge set $E$ is also divided into three parts, $E_1$, $E_2$, and $E_3$.
As a result, $\mathcal{E}_1$ and $E_1$ are sets of edges whose endpoints are not $v_i$ or $v_j$;
$\mathcal{E}_2$ and $E_2$ are edge sets with $v_i$ as an endpoint;
$\mathcal{E}_3$ and $E_3$ are edge sets with $v_j$ as an endpoint.

As for graph $\mathcal{G}'$ which yields from adding edge $(v_i, v_j)$ to $\mathcal{G}$, the split is completed in the same way: $\mathcal{E}'=\mathcal{E}_1\cup\mathcal{E}_2\cup\mathcal{E}_3\cup \mathcal{E}_{ij}$, $E'=E_1\cup E_2\cup E_3\cup E_{ij}$.
Among them, $\mathcal{E}_{ij}=\{(v_i,v_j)\}$ , $E_{ij}=\mathcal{E}_{ij}$ if $v_i$ and $v_j$ are of different classes and $E_{ij}=\varnothing$ if they are of the same class.

Owing to the split, the $\sum_{(i,j)}\hat{A}_{ij}$ in Eq. (\ref{EPR}) can be simplified.
To begin with, denote the normalized adjacency matrix of $\mathcal{G}'$, $\widetilde{D}'^{-1/2}\widetilde{A}'\widetilde{D}'^{-1/2}$ as $\hat{A}'$.
Furthermore, define
\begin{equation}
    \begin{aligned}
m_l=\sum_{(v_i,v_j)\in\mathcal{E}_l}\hat{A}_{ij},\;
e_l=\sum_{(v_i,v_j)\in E_l}\hat{A}_{ij},\;l=1,2,3,\nonumber
    \end{aligned}
\end{equation}
and
\begin{equation}
    \begin{aligned}
m'_l=\sum_{(v_i,v_j)\in\mathcal{E}_l}\hat{A}'_{ij},\;
e'_l=\sum_{(v_i,v_j)\in E_l}\hat{A}'_{ij},\;l=1,2,3.\nonumber
    \end{aligned}
\end{equation}
Thus, $r_\mathcal{G}$ and $r_{\mathcal{G}'}$ can be rewritten as
\begin{equation}
    r_{\mathcal{G}} = \frac{\sum_{i=1}^3e_i}{\sum_{i=1}^3m_i},\;r_{\mathcal{G}'} = \frac{\sum_{i=1}^3e'_i + \sum_{(v_i,v_j)\in E_{ij}}\hat{A}'_{ij}}{\sum_{i=1}^3m'_i + \sum_{(v_i,v_j)\in \mathcal{E}_{ij}}\hat{A}'_{ij}}.\nonumber
\end{equation}

Considering the change in node degrees in graph $\mathcal{G}$ after adding edge $(v_i, v_j)$ to it, it is easy to know that each node except $v_i$ and $v_j$ maintains its own degree while the degrees of $v_i$ and $v_j$ are increased by $1$, indicating that
{\footnotesize\begin{equation}
    \left\{\begin{aligned}
        m'_1 &=  m_1,\\
        m'_2 &= \sqrt{\frac{d_i}{d_i+1}}m_2,\\
        m'_3 &= \sqrt{\frac{d_j}{d_j+1}}m_3,\\
        \hat{A}'_{ij} &= \hat{A}'_{ji} = \frac{1}{\sqrt{(d_i+1)(d_j+1)}}.
        \nonumber
    \end{aligned}\right.    
\end{equation}}
As a result, $r_\mathcal{G}$ and $r_{\mathcal{G}'}$ can be formulated as
\begin{equation}
    \begin{aligned}
        r_\mathcal{G} &= \frac{e_1+e_2+e_3}{m_1+m_2+m_3},\\
        r_{\mathcal{G}'} &= \frac{e_1+\sqrt{\frac{d_i}{d_i+1}}e_2+\sqrt{\frac{d_j}{d_j+1}}e_3+\xi\cdot\frac{2}{\sqrt{(d_i+1)(d_j+1)}}}{m_1+\sqrt{\frac{d_i}{d_i+1}}m_2+\sqrt{\frac{d_j}{d_j+1}}m_3+\frac{2}{\sqrt{(d_i+1)(d_j+1)}}},
    \end{aligned}
\end{equation}
where $\xi=0$ if $v_i$ and $v_j$ are of the same class and $\xi=1$ otherwise.

Let $m = \sum_{i=1}^3m_i$, $m' = \sum_{i=1}^3m'_i+ \frac{2}{\sqrt{(d_i+1)(d_j+1)}}$, which are the sum of elements in $\hat{A}$ and $\hat{A}'$, respectively.
To compare $r_\mathcal{G}$ with $r_{\mathcal{G}'}$, we introduce
\begin{equation}
    \delta_{\mathcal{G}, \mathcal{G}'} = m\cdot m'\cdot (r_\mathcal{G}-r_{\mathcal{G}'}).
    \label{delta}
\end{equation}
As $\delta_{\mathcal{G},\mathcal{G}'}\propto(r_\mathcal{G}-r_{\mathcal{G}'})$, it can effectively reflects the trends of changes in EPR.
With a simple constraint, $\delta_{\mathcal{G},\mathcal{G}'}$ can be related to the edge added, which is formally summarzied as Theorem 3.3.
\paragraph{Theorem 3.3.} Given a graph $\mathcal{G}$ with $d_{max}\leq 4d_{min}-1$ and $d_{min}\geq 1$, then:
\begin{enumerate}
     \item If $\mathcal{G}'$ yields from adding an edge between two same-class nodes to $\mathcal{G}$ (i.e., $\xi=0$), then $\delta_{\mathcal{G},\mathcal{G}'}>0$.
     \item If $\mathcal{G}'$ yields from adding an edge between two different-class nodes to $\mathcal{G}$ (i.e., $\xi=1$), then $\delta_{\mathcal{G},\mathcal{G}'}<0$.
\end{enumerate}
The proof of Theorem 3.3 is in Appendix A.

Theorem 3.3 indicates that adding edge between same-class nodes decreases EPR of the graph, while adding edge between different-class nodes leads to the opposite result.
The conclusion also meets our expectation of EPR and edge adding as well.

Moreover, motivating by the proof of Theorem 3.3.2, an additional assumption is used to precisely quantify $\delta_{\mathcal{G},\mathcal{G}'}$.
With the assumption, we can further compare the effect of edge adding and dropping. 
\paragraph{Assumption 3.4.} For each node in the graph $\mathcal{G}$, constantly $k$ of all the message passed in is error message, i.e.
\begin{equation}
    M_{wp,i}/M_i=k\;,\;\forall i\in \{1,2,\cdots, N\},\nonumber
\end{equation}
where $M_{wp,i}$ is the amount of message wrongly passed to $v_i$, while $M_i$ is the amount of all the message passed to $v_i$.

With Assumption 3.4, it can be derived that
\begin{equation}
    \delta_{\mathcal{G},\mathcal{G}'} = (k-\xi)\cdot\alpha_{i,j}\cdot m,
    \label{add_edge}
\end{equation}
where $\alpha_{i,j} = 2/\sqrt{(d_i+1)(d_j+1)}$.
The detailed derivation is in Appendix B.

Now we apply the above analysis to edge dropping to answer the question raised in the title of this subsection.
Specifically, we may consider $\mathcal{G}'$ as the original graph and $\mathcal{G}$ as the graph augmented by dropping edge $(v_i, v_j)$.
Hence, for edge dropping, it holds that
\begin{equation}
    \delta_{\mathcal{G}',\mathcal{G}} = (\xi-k)\cdot\alpha_{i,j}\cdot m.
    \label{drop_edge}
\end{equation}
The detailed derivation is in Appendix C.
Hence, the following conclusions are drawn from Eq. (\ref{delta}), Eq. (\ref{add_edge}), and Eq. (\ref{drop_edge}):
\paragraph{Theorem 3.5.} Under Assumption 3.4, if graph $\mathcal{G}'$ is obtained through adding edge $(v_i,v_j)$ to $\mathcal{G}$, it holds that
\begin{equation}
    \left\{\begin{aligned}
        r_{\mathcal{G}'}-r_\mathcal{G}&=-\frac{k}{m'}\cdot\alpha_{i,j} &\text{if}\; c(v_i) = c(v_j),\\
        r_{\mathcal{G}'}-r_\mathcal{G}&=\frac{1-k}{m'}\cdot\alpha_{i,j} &\text{if}\; c(v_i) \neq c(v_j),
        \nonumber
    \end{aligned}\right.    
\end{equation}
where $c(v)$ denotes the class of node $v$.

For most graphs, the EPR is less than 0.5, i.e., $k<0.5$, which \textbf{\textit{explains why dropping edges is more stable than adding edges for most of the time}}.
What's more, as mentioned in Section 2.2, GCA tends to drop edges with low importance, which corresponds to a higher $\alpha_{i,j}$ and leads to a more unstable result according to Theorem 3.5.
\subsection{Adding Edges with Retaining EPR}
As shown in Section 3.3, adding edges is usually worse than dropping edges.
Since the edge dropping may not work when the graph is too sparse, a direct question is that \textit{whether edge adding can work like edge dropping}?
According to Theorem 3.5, for dropping edge, the change of EPR $\propto\alpha_{i,j}$, while for adding edges, the change of EPR $\propto\alpha_{i,j}/m'\approx\alpha_{i,j}/m\propto\alpha_{i,j}$.

Following the theoretical conclusion, we propose Error-PAssing-based Graph Contrastive Learning (EPAGCL), which generates views for GCL based on the $\alpha_{i,j}$ corresponding to each edge, ensuring that the EPR of the graph will not increase too much even if the edge is wrongly added or dropped.

To be specific, edges are added or dropped  by sampling from the corresponding probability \cite{GRACE}.
The edge set $\widetilde{\mathcal{E}}$ of the generated view can be formulated as $\widetilde{\mathcal{E}}=\widetilde{E}\cup\widetilde{E}'$, with probability
\begin{equation}
    P\{(v_i,v_j)\in \widetilde{E}\} = 1 - p_{ij}^d\label{drop}
\end{equation}
and
\begin{equation}
    P\{(v_i,v_j)\in \widetilde{E}'\} = p_{ij}^a,\label{add}
\end{equation}
where $\widetilde{E}$ is a subset of the original edge set $\mathcal{E}$ and $\widetilde{E}'$ is a subset of the to-be-added edge set $\mathcal{E}_a$.
$p_{ij}^a$ and $p_{ij}^d$ stand for the probability of dropping and adding $(v_i,v_j)$ respectively.
\begin{algorithm}[t]
\caption{Algorithm to select added edges}
\label{Algorithm 1}
\textbf{Input}: Vertex set $\mathcal{V}$, edge set $\mathcal{E}$.\\
\textbf{Output}: Drop-edge weight $w^d$, add-edge weight $w^a$, set of edges to be added $\mathcal{E}_a$.
    \begin{algorithmic}[1]
        \STATE $l\leftarrow$ $|\mathcal{E}|$
        \FOR{$e_{i,j}$ in $\mathcal{E}$}
            \STATE $w^d_{ij}\leftarrow \alpha_{i,j;\text{ drop}}$\,\qquad\qquad\quad\;\, $//$\;$\alpha_{i,j}$ of $(\mathcal{V},\mathcal{E})$ as $\mathcal{G}'$.

        \ENDFOR
        \STATE $\mathcal{V}_a\leftarrow$ vertex in $\mathcal{V}$ of top $\sqrt{2l}$ degrees
        \STATE $\mathcal{E}_a\leftarrow \mathcal{V}_a\times\mathcal{V}_a-\mathcal{E}$ \qquad\qquad\qquad \qquad$//$\;$l\leq$ $|\mathcal{E}_a|$ $\leq 2l$.
        \FOR{$e_{i,j}$ in $\mathcal{E}_a$}
        \STATE $w^a_{ij}\leftarrow \alpha_{i,j;\text{ add}}$\qquad\qquad\qquad\;\,$//$\;$\alpha_{i,j}$ of $(\mathcal{V},\mathcal{E})$ as $\mathcal{G}$.
        \ENDFOR
        \STATE \textbf{return} $w^d$, $w^a$, $\mathcal{E}_a$
    \end{algorithmic}
\end{algorithm}
Algorithm \ref{Algorithm 1} shows how to get the to-be-added edge set $\mathcal{E}_a$ along with the weights $w_d$ and $w_a$.
Note that for edges to be dropped, $\alpha_{i,j}=2/\sqrt{d_id_j}$, while for edges to be added, $\alpha_{i,j}=2/\sqrt{(d_{i}+1)(d_{j}+1)}$ (refer to Appendix C).
The weights are then transformed into probability through a normalization step \cite{GCA}.
\begin{equation}
    \left\{\begin{aligned}
        p_{ij}^a &= \min(\frac{\max(w^a)-w^a_{ij}}{\max(w^a)-\mu_{w^a}}\cdot p_{\text{add}},\; p_\tau),\\
        p_{ij}^d &= \min(\frac{w^d_{ij}-\min(w^d)}{\mu_{w^d}-\min(w^d)}\cdot p_{\text{drop}},\; p'_\tau).
    \end{aligned}\right.
    \label{weights}
\end{equation}
In Eq. (\ref{weights}), $p_{\text{add}}$ and $p_{\text{drop}}$ are hyper-parameters that controls the overall probability.
$p_\tau, p'_\tau$ are cut-off probability that is no greater than 1.
$\mu_{w^a}$ and $\mu_{w^d}$ stand for the average of $w^a$ and $w^d$, respectively.

Note that the weights is obtained based on the original graph, they will be computed only once, which adds almost nothing to the burden.
As for graphs with a significant number of nodes, thanks to the nodes filtering steps (line 5 in Algorithm \ref{Algorithm 1}), the computation is greatly accelerated and can be finished within an acceptable timeframe.

The training algorithm is summarized as pseudo-code in Algorithm \ref{Algorithm 2}.
As Theorem 3.5 shows, dropping edges is a more stable way to generate augmented views, so it is used when generating both the views, while adding edges is used only for generating one views.
Other than edge perturbation, random feature mask, which is widely used in graph presentation learning \cite{GRACE} \cite{MVGRL}, is also employed.
After the views are generated, an InfoNCE-like objective \cite{InfoNCE} is employed.
For each positive pair $(u_i,v_i)$ in $G_1,G_2$, which is the embedding corresponds the same node of the original graph, we define $s(u_i,v_i)$ as the cosine similarity of $g(u_i)$ and $g(v_i)$, where $g(\cdot)$ is a projection head \cite{MI}, and
\begin{equation}
    \begin{aligned}
        &l(u_i,v_i)=\\
        &\quad\log\frac{e^{s(u_i,v_i)/\tau}}{e^{s(u_i,v_i)/\tau}+\sum\limits_{i\neq j}e^{s(u_i,v_j)/\tau}+\sum\limits_{i\neq j}e^{s(u_i,u_j)/\tau}},
    \end{aligned}
\end{equation}
where $\tau$ is a temperature parameter.
The contrastive loss is then computed by added up $l(u_i,v_i)$ and $l(v_i,u_i)$ for all $i\in\{1,2,\cdots,N\}$.
\begin{algorithm}[t]
\caption{The EPAGCL training algorithm}
\label{Algorithm 2}
\textbf{Input}: Original graph $G=(V,E)$ with feature $X$, weights $w^a,w^d$, to-be-added edge set $\mathcal{E}_a$.
    \begin{algorithmic}[1]
        \FOR{epoch $\leftarrow 1,2,\cdots$}
        \STATE Obtain $p_{ij}^a,p_{ij}^d$ according to $w^a,w^d$ through Eq. (\ref{weights}).
        \STATE Obtain $E_1, E_2$ according to $p_{ij}^d, E$ through Eq. (\ref{drop}).
        \STATE Obtain $E'$ according to $p_{ij}^a, \mathcal{E}_a$ through Eq. (\ref{add}).
        \STATE Apply random mask on feature $X$ and get $X_1$, $X_2$.
        \STATE Two views are generated: $G_1=(V,E_1)$ with feature $X_1$ and $G_2=(V,E_2\cup E')$ with feature $X_2$.
        \STATE Compute the contrastive loss $\mathcal{L}$ between $G_1$ and $G_2$.
        \STATE Update the parameters to minimize $\mathcal{L}$.
        \ENDFOR
    \end{algorithmic}
\end{algorithm}
\section{Experiments}
In this section, we perform experiments to investigate the following questions:
\begin{itemize}
    \item[\textbf{Q1}] Does EPAGCL outperforms the existing baseline methods on node classification?
    \item[\textbf{Q2}] What is the time and memory burden of EPAGCL?
    \item[\textbf{Q3}] How does each part of the proposed augmentation strategy affect the effectiveness of training?
\end{itemize}
\begin{table}[ht]
    \centering
    \setlength{\tabcolsep}{1mm}
    \fontsize{9}{11}\selectfont{
    \begin{tabular}{c|cccc}
        \hline\toprule
        \textbf{Dataset} & \textbf{Nodes} & \textbf{Edges} & \textbf{Features} & \textbf{Classes} \\ 
        \hline
        Cora & 2,708 & 5,278 & 1,433 & 7 \\ 
        CiteSeer & 3,327 & 4,552 & 3,703 & 6 \\ 
        PubMed & 19,717 & 44,324 & 500 & 3 \\ 
        WikiCS & 11,701 & 216,123 & 300 & 10 \\ 
        Amazon-Photo & 7,650 & 119,081 & 745 & 8 \\ 
        Coauthor-Physics & 34,493 & 247,962 & 8,415 & 5 \\
        Ogbn-Arxiv & 169,343 & 1,166,243 & 128 & 40 \\
        \bottomrule\hline
    \end{tabular}
    }
    \caption{Statistics of datasets used in experiments.}
    \label{Datasets}
\end{table}
\begin{table*}[t]
    \centering
    \setlength{\tabcolsep}{2mm}
    \fontsize{9}{10}\selectfont{
    \begin{tabular}{c|ccccccc|c}
        \hline\toprule
        \textbf{Method} & \textbf{Cora} & \textbf{CiteSeer} & \textbf{PubMed} & \textbf{WikiCS} & \textbf{Amz. Photo} & \textbf{Co. Physics} & \textbf{ogbn-arxiv} & \textbf{Rank}\\ 
        \hline
        GCN & 84.15 $\pm$ 0.31 & 72.00 $\pm$ 0.39 & 85.39 $\pm$ 0.28 & 79.93 $\pm$ 0.53 & 92.69 $\pm$ 0.26 & 95.14 $\pm$ 0.41 & 69.44 $\pm$ 0.06 & -\\ 
        GAT & 84.08 $\pm$ 0.51 & 72.17 $\pm$ 0.39 & 84.91 $\pm$ 0.20 & 80.84 $\pm$ 0.21 & 92.39 $\pm$ 0.30 & 95.48 $\pm$ 0.14 & OOM & -\\ 
        \hline
        DGI & 82.47 $\pm$ 0.38 & 71.03 $\pm$ 0.88 & 85.64 $\pm$ 0.39 & 79.43 $\pm$ 0.27 & 91.00 $\pm$ 0.37 & OOM & OOM & 7.8\\ 
        GMI & 82.90 $\pm$ 0.69 & 69.51 $\pm$ 0.72 & 83.37 $\pm$ 0.52 & 80.23 $\pm$ 0.29 & 90.88 $\pm$ 0.31 & OOM & OOM & 8.6\\ 
        MVGRL & 81.46 $\pm$ 0.40 & 70.78 $\pm$ 0.53 & 84.33 $\pm$ 0.25 & 80.10 $\pm$ 0.26 & 90.91 $\pm$ 0.68 & 94.91 $\pm$ 0.16 & OOM & 8.2\\ 
        GRACE & 85.34 $\pm$ 0.29 & 71.66 $\pm$ 0.35 & \underline{86.74 $\pm$ 0.18} & 80.65 $\pm$ 0.20 & 93.13 $\pm$ 0.12 & 95.63 $\pm$ 0.05 & 68.49 $\pm$ 0.01 & 3.6\\ 
        GCA* & 84.55 $\pm$ 0.43 & 70.81 $\pm$ 0.57 & 86.48 $\pm$ 0.17 & \underline{81.45 $\pm$ 0.15} & 93.08 $\pm$ 0.18 & 95.20 $\pm$ 0.09 & \underline{69.26 $\pm$ 0.01} & 4.6\\ 
        GCA & \underline{85.59 $\pm$ 0.35} & 71.21 $\pm$ 0.55 & 86.60 $\pm$ 0.19 & 79.18 $\pm$ 0.34 & 93.19 $\pm$ 0.24 & 95.28 $\pm$ 0.19 & 69.18 $\pm$ 0.01 & 4.7\\ 
        BGRL & 84.34 $\pm$ 0.36 & 70.02 $\pm$ 0.70 & 85.88 $\pm$ 0.14 & 80.43 $\pm$ 0.47 & 92.78 $\pm$ 0.65 & 95.56 $\pm$ 0.07 & 68.80 $\pm$ 0.10 & 6.1\\ 
        GREET & 80.42 $\pm$ 0.25 & 71.48 $\pm$ 0.99 & 86.27 $\pm$ 0.32 & 79.91 $\pm$ 0.50 & \textbf{93.56 $\pm$ 0.14} & \textbf{96.06 $\pm$ 0.11} & OOM & 5.0\\ 
        \hline
        EPAGCL* & 85.04 $\pm$ 0.33 & \textbf{71.97 $\pm$ 0.62} & 86.72 $\pm$ 0.11 & \textbf{81.81 $\pm$ 0.18} & 93.05 $\pm$ 0.23 & 95.41 $\pm$ 0.03 & \textbf{69.29 $\pm$ 0.01} & \underline{3.0}\\ 
        EPAGCL & \textbf{86.07 $\pm$ 0.32} & \underline{71.94 $\pm$ 0.57} & \textbf{86.77 $\pm$ 0.14} & 81.19 $\pm$ 0.11 & \underline{93.42 $\pm$ 0.12} & \underline{95.87 $\pm$ 0.04} & 69.25 $\pm$ 0.01 & \textbf{2.0}\\
        \bottomrule\hline
    \end{tabular}
    }
    \caption{Results in terms of classification accuracy (in percent $\pm$ standard deviation) on seven datasets. * means that we do not employ feature mask to augment graph data. OOM indicates Out-Of-Memory on a 40GB GPU. The best and runner-up results of self-supervised methods on each dataset are highlighted with \textbf{bold} and \underline{underline}, respectively.}
    \label{Accuracy}
\end{table*}
\subsection{Datasets}

Seven benchmark graph datasets are utilized for experimental study, including three citation network \textbf{Cora}, \textbf{CiteSeer} and \textbf{PubMed} \cite{Planetoid}, a reference network \textbf{Wiki-CS} \cite{WikiCS}, a co-purchase network \textbf{Amazon-Photo} \cite{COs}, a co-authorship network \textbf{Coauthor-Physics} \cite{COs} and a large-scale citation network \textbf{ogbn-arxiv} \cite{arxiv}.
The details of the datasets are summarized in Table \ref{Datasets}.
\subsection{Experimental Settings}
\paragraph{Backbone} For our proposed method, a two-layer GCN network \cite{GCN} with PReLU activation is applied.
The dimension of the hidden layer is 512 and the dimension of the final embedding is set as 256.
Also, we employ a projection head, which consists of a 256-dimension fully connected layer with ReLU activation and a 256-dimension linear layer.
The hyper-parameters of the training vary for different datasets, the details of which are shown in Appendix D.
\paragraph{Baselines} We compare EPAGCL with two groups of baseline methods:
(1) semi-supervised learning methods (i.e., \textbf{GCN} \cite{GCN} and \textbf{GAT} \cite{GAT});
(2) contrastive learning methods (i.e., \textbf{DGI} \cite{DGI}, \textbf{GMI} \cite{GMI}, \textbf{MVGRL} \cite{MVGRL}, \textbf{GRACE} \cite{GRACE}, \textbf{GCA} \cite{GCA}, \textbf{BGRL} \cite{BGRL}, and \textbf{GREET} \cite{GREET}). 
\paragraph{Evaluation Settings} To evaluate the proposed method, we follow the standard linear evaluation scheme introduced in \cite{DGI}.
Firstly, each model is trained in an unsupervised manner.
The resulting embeddings are then utilized to train and test a simple $l_2$-regularized logistic regression classifier, which is initialized randomly and trained with an Adam SGD optimizer \cite{Adam} for 3000 epochs.
The learning rate and weight decay factor of the optimizer are fixed to 0.01 and 0.0 respectively.

For each method on each dataset, the accuracy is averaged over 5 runs.
For each run, the dataset is randomly split, where 10\%, 10\% and the rest 80\% of the nodes are selected for the training, validation and test set, respectively.

The experiments are conducted on an NVIDIA A100 GPU with 40 GB memory.
\begin{figure*}[t]
    \centering
    \subfloat[Raw features of Cora]{
        \includegraphics[width=0.24\linewidth]{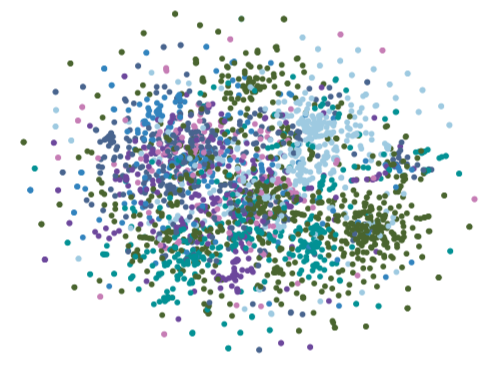}
    }
    \hfill
    \subfloat[EPAGCL on Cora features]{
        \includegraphics[width=0.24\linewidth]{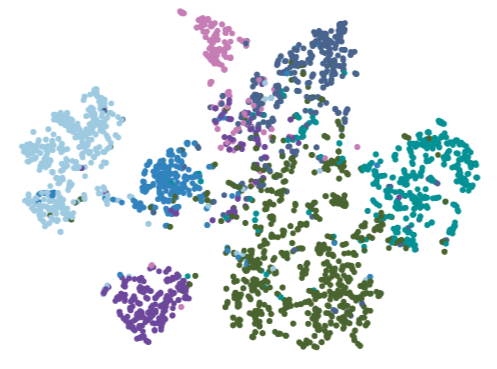}
    }
    \hfill
    \subfloat[Raw features of CiteSeer]{
        \includegraphics[width=0.24\linewidth]{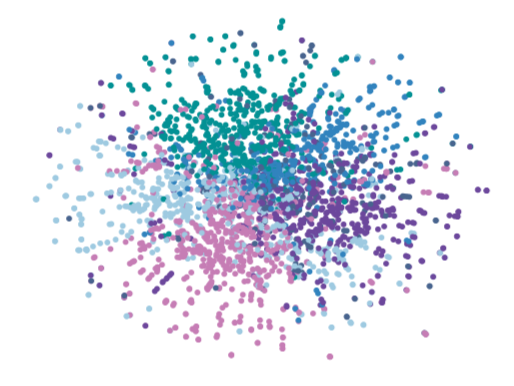}
    }
    \hfill
    \subfloat[EPAGCL on CiteSeer features]{
        \includegraphics[width=0.24\linewidth]{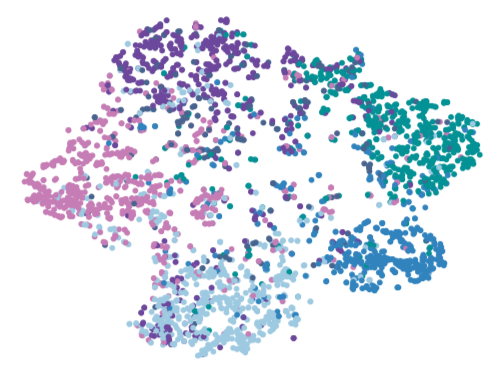}
    }
    \caption{t-SNE embeddings of the raw features and learned embeddings obtained through EPAGCL on Cora and CiteSeer.}
    \label{visualization}
\end{figure*}
\begin{table*}[t]
    \centering
    \setlength{\tabcolsep}{2mm}
    \fontsize{9}{10}\selectfont{
    \begin{tabular}{ccccc|cccc|cccc}
        \hline\toprule
        \multirow{2}{*}{\textbf{Method}} & \multicolumn{4}{c}{\textbf{Cora}} & \multicolumn{4}{c}{\textbf{PubMed}} & \multicolumn{4}{c}{\textbf{ogbn-arxiv}} \\
        \cmidrule(lr){2-5}\cmidrule(lr){6-9}\cmidrule(lr){10-13}
         & Proc. & FLOPs & Time & Mem. & Proc. & FLOPs & Time & Mem. & Proc. & FLOPs & Time & Mem. \\
        \midrule
        MVGRL & 0.18s & 5.55e6 & 0.03s & 3.49G & 15.88s & 4.04e7 & 0.16s & 21.25G & - & - & - & - \\
        GCA & 0.40s & 3.57e8 & 0.04s & 0.93G & 0.14s & 2.60e9 & 0.42s & 13.87G & 0.13s & 2.23e10 & 3.89s & 18.36G \\
        GREET & 4.46s & 5.90e9 & 0.15s & 1.47G & 240.01s & 2.18e10 & 6.38s & 39.50G & 15.95h & - & - & - \\
        EPAGCL & 0.40s & 3.57e8 & 0.04s & 0.90G & 0.57s & 2.60e9 & 0.42s & 17.10G & 21.94s & 2.23e10 & 3.89s & 22.93G \\
        \bottomrule\hline
    \end{tabular}
    }
    \caption{Comparison in terms of data pre-processing time, flops and training time of one epoch, and memory costs between different graph contrastive methods. Proc. and Mem. stand for data processing time and Memroy cost, respectively. `-' indicates Out-Of-Memory on a 40GB GPU. On ogbn-arxiv, the model is trained with a batch size of 256  due to the memory limit.}
    \label{burden}
\end{table*}
\subsection{Performance Analysis (Q1)} The classification accuracy is shown in Table \ref{Accuracy} with a comparative rank.
Specifically, despite the perturb target is different, our edge perturbation method is similar to GCA, while the feature mask is randomly applied.
Further experiments are conducted for comparison.
In the table, * means that we do not employ feature mask as one of the augmentations on graph data.

It's easy to find out that EPAGCL achieves better performance than baselines on almost every dataset.
For instance, on Cora, our method achieves a 0.48\% accuracy gain than the next best approach.
What's more, with the same contrastive objective, EPAGCL outperforms GCA by an average 0.61\% increase, which indicates that our augmentation method is more effective.
Thirdly, our method shows a lower standard deviation than GCA on most datasets, revealing its stability.
This also verifies the inference from Theorem 3.5 that dropping edges corresponding to higher $\alpha_{i,j}$ will lead to a more unstable result.

In additional, Figure \ref{visualization} displays t-SNE \cite{tSNE} plots of the raw feature and learned embeddings on Cora and CiteSeer.
\subsection{Efficiency Analysis (Q2)}
To illustrate the efficiency of our model, we compare our method with other graph contrastive methods in terms of data pre-processing time, flops and training time of one epoch, and memory costs.
MVGRL \cite{MVGRL} is a method that makes use of graph diffusion for contrast, which is kind of similar to edge adding.
GCA \cite{GCA} calculates different drop possibilities for edge dropping.
And GREET \cite{GREET} extracts information from feature and structure to benefit training.
Experiments are conducted on Cora, PubMed, and ogbn-arxiv, corresponding to small, big and huge datasets.
On ogbn-arxiv, the model is trained with a batch size of 256 because of the memory limit.
The results are shown in Table \ref{burden}.

The experiments show that although our methods takes more time to pre-process data and more memory for training than GCA, the additional burden is relatively small even when it is applied on a huge dataset like ogbn-arxiv.
As for MVGRL and GREET, the pre-processing time and memory requirement are much higher.
Specifically, on ogbn-arxiv, MVGRL reports Out-Of-Memory during the pre-processing phase.
While GREET takes a lot of time to pre-process data and finally reports OOM when training in regardless of batch size.
Moreover, the pre-processing time of EPAGCL grows in a slow rate with the increase of the scales of graph, which further indicates that EPAGCL also fits the huge datasets.
What's more, EPAGCL also shows a relatively fast training speed, especially compared with GREET.
\subsection{Ablation Study (Q3)}
Further experiments are conducted to demonstrate the effect of augment strategies.
We investigate the performance of the following six augmentations without feature mask on some benchmark datasets:
randomly add and drop edges as `Random Add'; add and drop edges adaptively on both views as `Add to Both Views'; our method as `EPAGCL'; drop edges adaptively as `Drop Only'; drop edges randomly as `Random Drop'; and add edges adaptively as `Add Only'.

\begin{figure}[t]
    \centering
    \includegraphics[width=\linewidth]{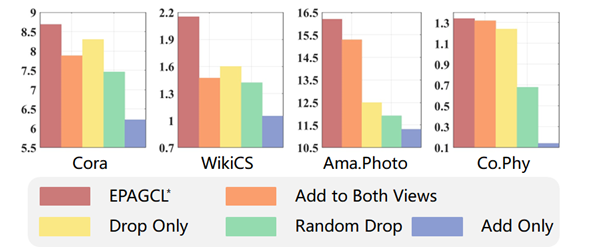}
    \caption{Performance improvement of five augmentation strategies compared to `Random Add'.}
    \label{Ablation}
\end{figure}

The results are shown in Figure \ref{Ablation}.
To better manifest the difference between the performance, we illustrate the performance improvement of the five augmentations compared with `Random Add'.
It can be observed that our method achieves the best performance on each dataset.
Moreover, `Drop Only' has an advantage over `Add Only'.
This indicates that adding edges is a relatively poorer augmentation way, which is consistent with our theory.
Thirdly, the adaptive strategy `Add to Both Views' leads to a great accuracy increase compared with the random strategy `Random Add', which proves the effectiveness of our method.

To sum up, our adaptive augmentation strategy is effective for edge-dropping and edge-adding as well.
And adding edges to only one view fully utilizes the augmentation.
\section{Conclusion}
In this paper, we propose a novel algorithm EPAGCL for GCL.
The main idea of EPAGCL is to use adding edges as one of the augmentations to generate views.
We firstly introduce Error-Passing Rate (EPR) to measure the quality of a view.
Based on EPR, the magnitude of effect of edge perturbation is quantified.
Thus, we are able to add edges adaptively with low time and memory burden and without labels, which is also valid for edge dropping.
Extensive experiments validate the correctness of our theoretical and reveal the effectiveness of our algorithm.
\bibliography{aaai25}
\appendix
\newpage\null\newpage
\section{Proof of Theorem 3.3}\label{Proof}
\paragraph{Theorem 3.3.} Given a graph $\mathcal{G}$ with $d_{max}\leq 4d_{min}-1$ and $d_{min}\geq 1$, then:
\begin{enumerate}
     \item If $\mathcal{G}'$ yields from adding an edge between two same-class nodes to $\mathcal{G}$ (i.e., $\xi=0$), then $\delta_{\mathcal{G},\mathcal{G}'}>0$.
     \item If $\mathcal{G}'$ yields from adding an edge between two different-class nodes to $\mathcal{G}$ (i.e., $\xi=1$), then $\delta_{\mathcal{G},\mathcal{G}'}<0$.
\end{enumerate}
\paragraph{Proof for Theorem 3.3.1}As mentioned in section 3.2,
\begin{equation}
    \begin{aligned}
        r_G &= \frac{e_1+e_2+e_3}{m_1+m_2+m_3},\\
        r_{G'} &= \frac{e_1+\sqrt{\frac{d_i}{d_i+1}}e_2+\sqrt{\frac{d_j}{d_j+1}}e_3+\xi\cdot\frac{2}{\sqrt{(d_i+1)(d_j+1)}}}{m_1+\sqrt{\frac{d_i}{d_i+1}}m_2+\sqrt{\frac{d_j}{d_j+1}}m_3+\frac{2}{\sqrt{(d_i+1)(d_j+1)}}},\\
        \delta_{\mathcal{G}, \mathcal{G}'} &= \sum\limits_{i=1}^3m_i\cdot (\sum\limits_{i=1}^3m'_i+ \frac{2}{\sqrt{(d_i+1)(d_j+1)}})\\
        &\qquad\qquad\qquad\qquad\qquad\qquad\qquad\cdot (r_\mathcal{G}-r_{\mathcal{G}'}).
    \label{delta_appendix}
    \end{aligned}
\end{equation}
For $\xi=0$, we have
\begin{equation}
    \begin{aligned}
        &\delta_{\mathcal{G}, \mathcal{G}'}=\\
        &((k_i-1)m_2+(k_j-1)m_3+\frac{2}{\sqrt{(d_i+1)(d_j+1)}})e_1\\
        +&((1-k_i)m_1+(k_j-k_i)m_3+\frac{2}{\sqrt{(d_i+1)(d_j+1)}})e_2\\
        +&((1-k_j)m_1+(k_i-k_j)m_2+\frac{2}{\sqrt{(d_i+1)(d_j+1)}})e_3,\nonumber
    \end{aligned}
\end{equation}

where $k_i = \sqrt{\frac{d_i}{d_i+1}}$ and $k_j = \sqrt{\frac{d_j}{d_j+1}}$.

Considering that for $l = 1,2,3$, $e_l=\sum_{(v_i,v_j)\in E_l}\hat{A}_{ij}$ , $m_l=\sum_{(v_i,v_j)\in\mathcal{E}_l}\hat{A}_{ij}$and $\hat{A}_{ij}$ is positive for any $i,j$, we have
\begin{equation}
    e_l>0,m_l>0,\qquad l=1,2,3.\label{m}
\end{equation}
Thus, Theorem 3.3.1 can be derived from the following lemma:
\paragraph{Lemma A.1.}
\begin{equation}
    \left\{\begin{array}{rl}
         (\sqrt{\frac{d_i}{d_i+1}} - 1)m_2 & +(\sqrt{\frac{d_j}{d_j+1}} - 1)m_3 \\
         & + \frac{2}{\sqrt{(d_i+1)(d_j+1)}}> 0, \quad(*)\\
         (1 - \sqrt{\frac{d_i}{d_i+1}})m_1 & +  (\sqrt{\frac{d_j}{d_j+1}} - \sqrt{\frac{d_i}{d_i+1}})m_3 \\
         & + \frac{2}{\sqrt{(d_i+1)(d_j+1)}}>0, \quad(**) \\
         (1-\sqrt{\frac{d_j}{d_j+1}})m_1 & +(\sqrt{\frac{d_i}{d_i+1}} - \sqrt{\frac{d_j}{d_j+1}})m_2 \\
         & + \frac{2}{\sqrt{(d_i+1)(d_j+1)}}>0.\quad(***)
    \end{array}\nonumber
    \right.
\end{equation}
\paragraph{Proof for Lemma A.1}
According to the definition of $d_{min}$ and $d_{max}$, 
\begin{equation}
    \left\{\begin{array}{ll}
         m_1 > 0, & (a)  \\
         m_2\leq\frac{d_i}{\sqrt{d_id_{min}}} = \sqrt{\frac{d_i}{d_{min}}}, & (b) \\
         m_3\leq\sqrt{\frac{d_j}{d_{min}}}. & (c)
    \end{array}
    \right.\label{dmindmax}
\end{equation}
What's more, because of the symmetry, assumed that $d_i \geq d_j$, then
\begin{equation}
    1 > \sqrt{\frac{d_i}{d_i+1}}\geq \sqrt{\frac{d_j}{d_j+1}}.\label{assume}
\end{equation}
From Eq. (\ref{m}) and Eq. (\ref{assume}), we have
\begin{equation}
    (***)L.H.S. > \frac{2}{\sqrt{(d_i+1)(d_j+1)}}>0=(***)R.H.S.\nonumber
\end{equation}
For (*), from Eq. (\ref{dmindmax}) (b), Eq. (\ref{dmindmax}) (c) and Eq. (\ref{assume}), it can be derived that $(*)L.H.S.\geq (\sqrt{\frac{d_i}{d_i+1}}-1)\cdot\sqrt{\frac{d_i}{d_{min}}}+(\sqrt{\frac{d_j}{d_j+1}}-1)\cdot\sqrt{\frac{d_j}{d_{min}}}+\frac{2}{\sqrt{(d_i+1)(d_j+1)}}\overset{\triangle}{=}f(d_i,d_j)$.
To prove that (*) is valid, it is only necessary to prove that $f(d_i, d_j)>0$.

Note that
\begin{equation}
    \begin{aligned}
        \frac{\partial f}{\partial d_i} & =\frac{1}{2\sqrt{d_{min}}}(\frac{d_i+2}{(d_i+1)^{3/2}}-\frac{1}{\sqrt{d_i}}) \\
        &\qquad\qquad-\frac{1}{\sqrt{d_j+1}(d_i+1)^{3/2}} \\
        & =(d_i+1)^{-3/2}\frac{1}{2\sqrt{d_{min}}}\cdot \\
        &\qquad\qquad\left((d_i+2-\sqrt{\frac{(d_i+1)^3}{d_i}})-\frac{2\sqrt{d_{min}}}{\sqrt{d_j+1}}\right) \\
        & \overset{\triangle}{=}(d_i+1)^{-3/2}\cdot\frac{1}{2\sqrt{d_{min}}}\cdot g(d_i,d_j),\nonumber
    \end{aligned}    
\end{equation}
where $g(d_i,d_j)=d_i+2-\sqrt{\frac{(d_i+1)^3}{d_i}}-\frac{2\sqrt{d_{min}}}{\sqrt{d_j+1}}$.

Furthermore,
\begin{equation}
    \frac{\partial g}{\partial d_i} = 1-\frac{\sqrt{d_i+1}(2d_i-1)}{2d_i^{3/2}} = 1 - \sqrt{\frac{4d_i^3-3d_i+1}{4d_i^3}}.\nonumber
\end{equation}

As $d_i\geq d_{min}\geq 1$ (assumption in Theroem 3.3), it holds that $\frac{\partial g}{\partial d_i} > 0$.
Thus, $g(d_i,d_j)$ is monotonically increasing with respect to $d_i$.
What's more, with $\frac{\partial g}{\partial d_j} = \frac{\sqrt{d_{min}}}{\sqrt{(d_j+1)^3}} > 0$, $g(d_i,d_j)$ is also monotonically increasing with respect to $d_j$.
Considering that $d_i, d_j\leq d_{max}\leq 4d_{min}-1$, we have
\begin{equation}
    \begin{aligned}
        g(d_i,d_j)& \leq g(d_{max}, d_{max})\\
        & = d_{max}+2-\sqrt{\frac{(d_{max}+1)^3}{d_{max}}}-\frac{2\sqrt{d_{min}}}{\sqrt{d_{max}+1}} \\
        & \leq d_{max}+2-\sqrt{\frac{(d_{max}+1)^3}{d_{max}}} -\frac{2\sqrt{d_{min}}}{\sqrt{4d_{min}}} \\
        & \leq d_{max}+2-1-\sqrt{\frac{(d_{max}+1)^3}{d_{max}}} \\
        & = (d_{max}+1)(1-\sqrt{\frac{d_{max}+1}{d_{max}}})<0,\nonumber
    \end{aligned}        
\end{equation}
which means that $\frac{\partial f}{\partial d_i} < 0\;,\;\forall d_i, d_j\in[d_{min}, d_{max}].$
Therefore, $f(d_i,d_j)$ is monotonically decreasing with respect to $d_i$.
Due to the symmetry, this also holds for $d_j$.
As a result, $f(d_i, d_j)\geq f(d_{max}, d_{max})$.
Hence, 
\begin{equation}
    \begin{aligned}
        & f(d_i,d_j)>0 \Leftrightarrow f(d_{max}, d_{max})>0\\
        \Leftrightarrow\quad & 2(\sqrt{\frac{d_{max}}{d_{max}+1}}-1)\sqrt{\frac{d_{max}}{d_{min}}}+\frac{2}{d_{max}+1} > 0 \\
        \Leftrightarrow\quad & \frac{1}{\sqrt{d_{max}+1}}>\sqrt{\frac{d_{max}}{d_{min}}}(\sqrt{d_{max}+1}-\sqrt{d_{max}}) \\
        \Leftrightarrow\quad & \sqrt{d_{min}}>\frac{\sqrt{d_{max}(d_{max}+1)}}{\sqrt{d_{max}+1}+\sqrt{d_{max}}} \\
        \Leftrightarrow\quad & d_{min}>\frac{d_{max}(d_{max}+1)}{(\sqrt{d_{max}+1}+\sqrt{d_{max}})^2}.\nonumber
    \end{aligned}
\end{equation}

As $(\sqrt{d_{max}+1}+\sqrt{d_{max}})^2\geq 4\sqrt{d_{max}+1}\cdot\sqrt{d_{max}}$, along with $d_{max}\leq 4d_{min}-1$, it holds that
\begin{equation}
    \begin{aligned}
        \frac{d_{max}(d_{max}+1)}{(\sqrt{d_{max}+1}+\sqrt{d_{max}})^2} & \leq \frac{1}{4}\sqrt{d_{max}(d_{max}+1)} \\
        & \leq \frac{1}{4}\sqrt{4d_{min}(4d_{min}-1)} \\
        & < d_{min}.\nonumber
    \end{aligned}
\end{equation}
Thus, we have $(*)L.H.S. \geq f(d_i, d_j) > 0$.

For (**), from Eq. (\ref{dmindmax}) (a), Eq. (\ref{dmindmax}) (c) and Eq. (\ref{assume}), it can be derived that $(**)L.H.S.> (\sqrt{\frac{d_j}{d_j+1}}-\sqrt{\frac{d_i}{d_i+1}})\sqrt{\frac{d_j}{d_{min}}}+\frac{2}{\sqrt{(d_i+1)(d_j+1)}}\overset{\triangle}{=}h(d_i,d_j)$.
To prove that (**) is valid, it is only necessary to prove that $h(d_i, d_j)>0$.

Note that
\begin{equation}
    \begin{aligned}
        \frac{\partial h}{\partial d_j} & = \frac{1}{\sqrt{d_{min}}}(\frac{d_j+2}{2(d_j+1)^{3/2}}-\sqrt{\frac{d_i}{d_i+1}}\frac{1}{2\sqrt{d_j}})\\
        &\qquad\qquad-\frac{1}{\sqrt{d_i+1}(d_j+1)^{3/2}} \\
        & \overset{\triangle}{=} (d_j+1)^{-3/2}\cdot\frac{1}{2\sqrt{d_{min}}}l(d_i,d_j),
        \nonumber
    \end{aligned}
\end{equation}
where $l(d_i,d_j)=d_j+2-\sqrt{\frac{d_i}{d_i+1}}\sqrt{\frac{(d_j+1)^3}{d_j}}-\frac{2\sqrt{d_{min}}}{\sqrt{d_i+1}}$.

Furthermore,
\begin{equation}
    \begin{aligned}
        \frac{\partial l}{\partial d_j} &= 1-\sqrt{\frac{d_i}{d_i+1}}\frac{2d_j-1}{2}\sqrt{\frac{d_j+1}{d_j^3}}\\
        & = 1-\sqrt{\frac{d_i}{d_i+1}}\cdot\sqrt{\frac{4d_j^3-3d_j+1}{4d_j^3}}\\
        & > 1-\sqrt{\frac{4d_j^3-3d_j+1}{4d_j^3}}.\nonumber
    \end{aligned}
\end{equation}

As $d_i\geq d_j\geq 1$, it can be derived that $\frac{\partial l}{\partial d_j} > 0$.
Thus, $l(d_i, d_j)$ is monotonically increasing with respect to $d_j$, which means that $l(d_i, d_j)\leq l(d_i, d_i) = 1-\frac{2\sqrt{d_{min}}}{\sqrt{d_i+1}}$.
Note that $d_i+1\leq d_{max}+1\leq 4d_{min}$, so $l(d_i, d_j)\leq 1-\sqrt{\frac{4d_{min}}{d_i+1}}\leq 0$.
Hence, $\frac{\partial h}{\partial d_j} \leq 0\;,\;\forall d_i,d_j\in[d_{min}, d_{max}]$.
Therefore, $h(d_i, d_j)$ is monotonically decreasing with respect to $d_j$, along with $d_i\geq d_j$, we have
\begin{equation}
    h(d_i, d_j)\geq h(d_i, d_i)=\frac{2}{d_i+1}>0,\nonumber
\end{equation}
which indicates that (**) is valid.

In summary, Lemma A.1 is proved, which means that Theorem 3.3.1 is valid.$\hfill\square$
\paragraph{Proof for Theorem 3.3.2}
To begin with, a common assumption as follows is introduced:
\paragraph{Assumption A.2.}
In the graph $\mathcal{G}$, at least one edge lies between two nodes that are of the same class.

For most graphs, the assumption holds for sure.
Otherwise, the structural property of the graph is meaningless.

Back to the proof, according to Eq. (\ref{delta_appendix}), with $\xi=1$, it holds that
\begin{equation}
    \begin{aligned}
        &\delta_{\mathcal{G}, \mathcal{G}'}=\\
        &((k_i-1)m_2+(k_j-1)m_3+\frac{2}{\sqrt{(d_i+1)(d_j+1)}})e_1\\
        +&((1-k_i)m_1+(k_j-k_i)m_3+\frac{2}{\sqrt{(d_i+1)(d_j+1)}})e_2\\
        +&((1-k_j)m_1+(k_i-k_j)m_2+\frac{2}{\sqrt{(d_i+1)(d_j+1)}})e_3\\
        +&\frac{2(m_1+m_2+m_3)}{\sqrt{(d_i+1)(d_j+1)}},\nonumber
    \end{aligned}
\end{equation}
where $k_i = \sqrt{\frac{d_i}{d_i+1}}$ and $k_j = \sqrt{\frac{d_j}{d_j+1}}$.

Let $t_l = e_l/m_l\;,\; l=1,2,3$.
As $e_l$ is the sum of $\hat{A}_{ij}$ for $(v_i,v_j)\in\mathcal{E}_l$, while $m_l$ is the sum of $\hat{A}_{ij}$ for $(v_i,v_j)\in E_l$, it can be recognized that $t_l\leq 1$ because $\mathcal{E}_l\subseteq E_l$.
Let $t = \max\{t_1,t_2,t_3\}$, with Lemma A.1, we have
\begin{equation}
    \begin{aligned}
        &\delta_{\mathcal{G}, \mathcal{G}'}\leq\\
        &((k_i-1)m_2+(k_j-1)m_3+\frac{2}{\sqrt{(d_i+1)(d_j+1)}})tm_1\\
        +&((1-k_i)m_1+(k_j-k_i)m_3+\frac{2}{\sqrt{(d_i+1)(d_j+1)}})tm_2\\
        +&((1-k_j)m_1+(k_i-k_j)m_2+\frac{2}{\sqrt{(d_i+1)(d_j+1)}})tm_3\\
        +&\frac{2(m_1+m_2+m_3)}{\sqrt{(d_i+1)(d_j+1)}} = \frac{2(t-1)(m_1+m_2+m_3)}{\sqrt{(d_i+1)(d_j+1)}}\leq 0.\nonumber
    \end{aligned}
\end{equation}

The equal sign holds if and only if $t_1=t_2=t_3=t=1$, which means that all the edges in graph $\mathcal{G}$ is between nodes of different classes, conflicting with Assumption A.2.
As a result, Theorem 3.3.2 is valid.

To sum up, Theorem 3.3 is proved.$\hfill\square$
\section{Derivation of Eq. (5) in Section 3.3 from Assumption 3.4}
\paragraph{Assumption 3.4.}
\begin{equation}
    M_{wp,i}/M_i=k\;,\;\forall i\in \{1,2,\cdots, N\}.\nonumber
\end{equation}
\paragraph{Eq. (5)} 
\begin{equation}
    \delta_{\mathcal{G},\mathcal{G}'} = (k-s)\cdot\alpha_{i,j}\cdot m.\nonumber
\end{equation}
To begin with, with Assumption 3.4, we have
\begin{equation}
    \left\{\begin{aligned}
        e_1 & = k\cdot m_1,\\
        e_2 & = k\cdot m_2,\\
        e_3 & = k\cdot m_3.\\
    \end{aligned}\right.\nonumber
\end{equation}
Thus, 
\begin{equation}
    \begin{aligned}
        e'_2 &= \sqrt{\frac{d_i}{d_i+1}}e_2 = \sqrt{\frac{d_i}{d_i+1}}km_2 = km'_2,\\
        e'_3 &= \sqrt{\frac{d_j}{d_j+1}}e_3 = \sqrt{\frac{d_j}{d_j+1}}km_3 = km'_3,\\
        \text{and }e'_1 &= e_1 = (e_1+e_2+e_3)-(e_2+e_3)\\
             &= km-ke_2-ke_3 = km_1 = km'_1.\\
    \end{aligned}\nonumber
\end{equation}
So Eq. (\ref{delta_appendix}) can be formulated as
\begin{equation}
    \begin{aligned}
        \delta_{\mathcal{G},\mathcal{G}'} &= (e_1+e_2+e_3)(m'_1+m'_2+m'_3+\alpha_{i,j})\\
        &\qquad - (e'_1+e'_2+e'_3+s\alpha_{i,j})(m_1+m_2+m_3)\\
        &= km(m'_1+m'_2+m'_3+\alpha_{i,j})\\
        &\qquad -(km'_1+km'_2+km'_3+s\alpha_{i,j})m\\
        &= (k-s)\cdot\alpha_{i,j}\cdot m,\nonumber
    \end{aligned}
\end{equation}
where $\alpha_{i,j} = 2/\sqrt{(d_i+1)(d_j+1)}$.

\section{Derivation of Eq. (6) in Section 3.3}
\paragraph{Eq. (6)}
\begin{equation}
    \delta_{\mathcal{G}',\mathcal{G}} = (s-k)\cdot\alpha_{i,j}\cdot m.\nonumber
\end{equation}
As mentioned in Section 3.3, we may consider $\mathcal{G}'$ as the original graph and $\mathcal{G}$ as the graph augmented by dropping edge $(v_i, v_j)$.
Thus, it holds that
\begin{equation}
    \begin{aligned}
        \delta_{\mathcal{G}',\mathcal{G}} & = m'\cdot m\cdot (r_{\mathcal{G}'}-r_\mathcal{G}) = -m\cdot m'\cdot (r_{\mathcal{G}}-r_{\mathcal{G}'})\\
        & = -\delta_{\mathcal{G},\mathcal{G}'} =  (s-k)\cdot\alpha_{i,j}\cdot m.\nonumber
    \end{aligned}
\end{equation}
Other than the derivation, note that
\begin{equation}
    \alpha_{i,j} = 2/\sqrt{(d_i+1)(d_j+1)} = 2/\sqrt{d_i'd_j'},\nonumber
\end{equation}
where $d_i'$ and $d_j'$ are the degrees of $v_i$ and $v_j$ in the original graph $\mathcal{G}'$, respectively.
This explains why $p_{ij}^{\text{drop}}\propto 2/\sqrt{d_id_j}$ while $p_{ij}^{\text{add}}\propto 2/\sqrt{(d_i+1)(d_j+1)}$.
\section{Hyper-parameters for different datasets}\label{parameters}
\begin{table}[h]
    \centering
    \setlength{\tabcolsep}{1mm}
    \fontsize{10}{12}\selectfont{
    \begin{tabular}{c|cccccccc}
        \hline\toprule
        Dataset & $\mu$ & $\lambda$ & $p_{e1}$ & $p_{e2}$ & $p_{f1}$ & $p_{f2}$ & $t$ & $e$ \\
        \hline
        Cora & 1e-3 & 1e-4 & 0.2 & 0.3 & 0.1 & 0.1 & 0.3 & 500 \\
        CiteSeer & 1e-3 & 1e-4 & 0.2 & 0.3 & 0.1 & 0.1 & 0.3 & 500 \\
        PubMed & 1e-3 & 1e-4 & 0.2 & 0.3 & 0.1 & 0.1 & 0.3 & 1000 \\
        WikiCS & 1e-3 & 1e-4 & 0.2 & 0.3 & 0.1 & 0.1 & 0.3 & 3000 \\
        Amz.Photo & 0.01 & 1e-3 & 0.3 & 0.5 & 0.1 & 0.1 & 0.3 & 1000 \\
        Co.Physics & 0.01 & 1e-3 & 0.1 & 0.4 & 0.4 & 0.1 & 0.5 & 1000 \\
        ogbn-arxiv & 1e-3 & 1e-4 & 0.6 & 0.6 & 0.1 & 0.1 & 0.3 & 500 \\
        \bottomrule\hline
    \end{tabular}
    \caption{Hyper-parameters that vary for different datasets.}
    \label{Hyper-parameters}
    }
\end{table}
Some hyper-parameters of the experiment vary on different datasets, which is shown in Table \ref{Hyper-parameters}.
Specifically, we carry out grid search for the hyper-parameters on the following search space:
\begin{itemize}
    \item Learning rate for training $\mu$: $\{0.01, 1\text{e}-3, 1\text{e}-4\}$
    \item Weight decay for training $\lambda$: $\{1\text{e}-3, 1\text{e}-4\}$
    \item Edge dropping rates $p_{e1},p_{e2}$: $\{0.1,0.2,0.3,0.4,0.5,0.6\}$
    \item Feature masking rates $p_{f1}, p_{f2}$: $\{0.1, 0.3, 0.5\}$
    \item Temperature $t$: $\{0.3,0.4,0.5\}$
    \item Number of training epochs $e$: $\{500,1000,3000\}$.
\end{itemize}
For each dataset, a set of hyper-parameters is chosen to obtain the best average accuracy.
\section{Statistical Tests of Significance}
To demonstrate the significance of the improvement in performance, Wilcoxon signed-rank \cite{Wilcoxon} is used on Cora, CiteSeer, PubMed \cite{Planetoid}, and WikiCS \cite{WikiCS} as examples.
Specifically, we conduct Wilcoxon signed-rank test to compare the accuracy of EPAGCL with that of other methods.
For each method on each dataset, we test the following hypotheses at the significance level of $\alpha = 0.05$:
\begin{equation}
    H_1 \text{: The observations } X_i-Y_i \text{ are symmetric about } \mu > 0,\nonumber
\end{equation}
where $X_i$ is the accuracy of EPAGCL, and $Y_i$ is that of other methods.
The results are summarized in table \ref{wilcoxon1}.
\begin{table}[h]
    \centering
    \setlength{\tabcolsep}{2mm}
    \fontsize{10}{12}\selectfont{
    \begin{tabular}{c|cccc}
        \hline\toprule
        \textbf{Method} & \textbf{Cora} & \textbf{CiteSeer} & \textbf{PubMed} & \textbf{WikiCS} \\
        \midrule
        DGI & 2.98e-8 & 4.27e-5 & 2.98e-8 & 2.98e-8 \\
        GMI & 2.98e-8 & 2.98e-8 & 2.98e-8 & 2.98e-8 \\
        MVGRL & 2.98e-8 & 4.08e-6 & 2.98e-8 & 2.98e-8 \\
        GRACE & 2.45e-4 & 1.39e-3 & 2.58e-2 & 2.98e-8 \\
        GCA & 2.45e-4 & 2.45e-4 & 3.69e-3 & 2.98e-8 \\
        BGRL & 2.98e-8 & 2.98e-8 & 2.98e-8 & 2.98e-8 \\
        GREET & 2.98e-8 & 3.69e-3 & 2.98e-8 & 2.98e-8 \\
        \bottomrule\hline
    \end{tabular}
    }
    \caption{Probability of rejection of the hypothesis $H_1$. As we only have 5 runs for each dataset, the result of each run is repeated 5 times to obtain a size-of-25 observation.}
    \label{wilcoxon1}
\end{table}

As a result, it shows that our method has a significant improvement over others.
\section{Error Passing Rate of Benchmark Datasets}
To validate our claim in Section 3.3 that `for most graphs, the EPR is less than 0.5', further experiments are conducted. The results are listed in Table \ref{EPR_appendix}.
\begin{table}[h]
    \centering
    \setlength{\tabcolsep}{2mm}
    \fontsize{10}{12}\selectfont{
    \begin{tabular}{cc|cc}
        \hline\toprule
        \textbf{Dataset} & \textbf{EPR} & \textbf{Dataset} & \textbf{EPR}\\
        \midrule
        Cora & 0.168 & Co.CS & 0.160\\
        CiteSeer & 0.286 & Co.Phy & 0.081\\
        PubMed & 0.207 & Ama.computers & 0.171\\
        WikiCS & 0.288 & Ama.Photo & 0.132\\
        ogbn-arxiv & 0.352\\
        \bottomrule\hline
    \end{tabular}
    }
    \caption{EPR of real-world datasets.}
    \label{EPR_appendix}
\end{table}
\section{Analysis on Heterogeneous Graphs}
In the main body of the paper, only homogeneous graphs are discussed.
In fact, EPR is also a reasonable metric for heterogeneous graphs. To achieve the heterogeneous classification, heterogeneous GNNs require high-order neighbors. The bridging nodes make sense only if the receptive field is large enough. If the receptive field is just 1, the bridging nodes are still noisy neighbors.

Note that EPR is defined to quantify impact of 1-hop neighbors, so it is fair to regard bridging nodes as noise. We may define high-order EPR for better explanations on heterogeneous graphs.

Some experiments are conducted for validation.
We compare EPAGCL with GCA using the same hyper-parameters and a random 10\%/10\%/80\% split on Texas, Cornell, and Wisconsin.
The average results are listed in Table \ref{heterophily}, demonstrating that our augmentation method is effective on heterogeneous graphs.
\begin{table}[h]
    \centering
    \begin{tabular}{c|ccc}
        \hline\toprule
        \textbf{Method} & \textbf{Texas} & \textbf{Cornell} & \textbf{Wisconsin}\\
        \midrule
        GCA & 62.18 & 48.98 & 54.43\\
        EPAGCL & 64.08 & 51.74 & 57.61\\
        \bottomrule\hline
    \end{tabular}
    \caption{Results in terms of average classification accuracy over 5 runs on three heterogeneous datasets.}
    \label{heterophily}
\end{table}
\end{document}